\documentclass{article}
\usepackage{lineno,hyperref}
\modulolinenumbers[5]
\usepackage{amsmath}

\usepackage{systeme}
\usepackage{graphicx}
\usepackage{float}

\begin{document}
  

\title{A Framework for  Moment Invariants}

\author{Omar TAHRI }
%
%
\maketitle

\begin{abstract}
For more than half a century, moments have attracted lot ot interest in the pattern recognition community.
The moments of a distribution (an object)  provide several of its characteristics as center of gravity, orientation, disparity, volume. Moments can be used  to define invariant characteristics   to some transformations that an object 
can undergo, commonly called moment invariants.  This work provides a simple and systematic formalism  to 
compute geometric moment invariants in n-dimensional space.

\end{abstract}

\section*{keywords:}
Geometric moments, affine invariants, n-dimensional space.



\section{Introduction}
Since they were introduced in 1962 by Hu \cite{hu1962visual}, Moment Invariants
 have generated a lot of interest in the   pattern recognition field.  During 
 more that half a century, many theoretical frameworks and  applications   have been developed: 
 pose estimation \cite{cyganski1985applications,mukundan1998moment}, character recognition \cite{wong1995generation}, target recognition \cite{liu2012target}, quality inspection \cite{sluzek1995identification}, image matching \cite{chen2010zernike},  multi-sensors fusion \cite{markandey1992robot} and visual servoing \cite{tahri2005point}, \cite{tahri2010decoupled}.
 
Initially, moment invariants have  been mainly applied   in 2D space.  Subsequently,  invariant from moments defined in higher dimensional space have been defined \cite{markandey1992robot, tahri2010decoupled, dominguez2017simultaneous}. 
Although using moments of order higher  than 2 is not new \cite{canterakis3d,sadjadi1980three,novotni20033d}, such applications are  gaining interest thanks to  3D vision sensors.  

The first important result for deriving moment invariants is the fundamental theorem of moment invariants (FTMI) \cite{hu1962visual}. Hu employed his theorem to derive seven 2D moment invariants.  In fact, the FTMI contains some mistakes that have been emphasized by Mamistvalov in \cite{mamistvalov1970fundamental} in 1970 (in Russian). Despite this, the FTMI as proposed by Hu has been quoted in several works until $1991$  when  Reiss  established  the revised fundamental theorem of moment invariants (RFTMI) in 2D Space.  The FTMI was generalized for the n-dimensional case and applied in multi-sensor fusion in \cite{markandey1992robot}.  Unfortunately, the generalized theorem contains the same mistake as the one  given by Hu. Finally, Mamistvalov proposed  the correct generalization of the RFTMI to n-dimensional solids. The idea behind using the FTMI is that 
invariants of n-ary forms are also invariant in the case where the coefficients of the n-ary form are replaced by the corresponding moments.   \cite{kung1984invariant} proposed an algorithm for determining invariants of binary forms.  In \cite{Tahri03b}, a systematic method to derive independent moment invariants to orthogonal transformations in n-dimensional space has been  proposed.  The proposed scheme  is based on rotation speed tensor rather than rotation matrix to obtain the invariant to rotations. More recent works in the last decade proposed systematic schemes for 2D and 3D cases \cite{suk2011tensor,suk2011affine}, \cite{xu2008geometric}.  Despite the difference of terminology, the way to construct invariants in \cite{kung1984invariant} and in \cite{suk2011tensor} can be considered as similar. 

This paper proposes a unified and novel scheme to derive  affine invariants from geometric moments of n-dimensional solids. The idea behind the method is to consider that any affine transformation can be decomposed using SVD into a rotation followed by a non-uniform scale change then by another rotation.  Firstly, the method proposed in this paper is quite simple  to understand. Second, the method ensures that all possible invariants are obtained and  redundancy between different invariants   is quite easy to express and to eliminate using matrix algebra.  In the next section, basic definitions of moment in n-dimensional space are presented. Then, in the subsequent sections, invariance to scale changes, to rotation and  affine transformations will be dealt with.

 \section{Geometrical moments in n-dimensional space}
  \subsection{Notations}
 The following notations will be used in the sequel:
 \begin{itemize}
 	\item $n$: the dimension of the considered space.
 	\item $ m_{p_1 \hdots p_n} $ and $ \mu_{p_1 \hdots p_n} $ a moment and a centered moment of order $p=p_1+ \hdots + p_n$ of n-dimensional object.
 	\item ${\bf v}^1_p$:  vector composed by all moments of order $p$.
 	\item ${\bf v}^k_p$:  vector composed by all monomials   of degree $k$ and using as variable  the entries of ${\bf v}^1_p$.
 	\item ${\bf v}^{(kk'k''...)}_{(pp'p"...)}$: vector obtained by all possible products between the entries of ${\bf v}^k_p$, ${\bf v}^{k'}_{p'}$,....
 	\item ${\bf S}=diag(\sigma_1,\,\,\sigma_2,\hdots,\,\, \sigma_n)$:  scale change transformation in n-dimensional space
 	\item ${\bf R}$: orthogonal transformation in n-dimensional space.
 	\item ${\bf A}$: non-singular $n\times n$ non singular matrix  defining an affine transformation  in n-dimensional space.
    \item $i_s$: invariant to scale change.
    \item $i_r$: invariant to rotations.
    \item $i_a$: affine invariant.
 \end{itemize}

  \subsection{Geometrical moments in n-dimensional space}
 We first recall some basic definitions of moment functions.  Denoting ${\bf X} = (x_1, \hdots , x_n)$ the coordinates of a point in a n-dimensional space, the moments of the density function $f({\bf X})$ are defined by: 
 \begin{equation}
 \label{MomentsND}
 \begin{split}
 m_{p_1 \hdots p_n}  &=  \int_{-\infty }^{\infty } \hdots \int_{-\infty }^{\infty } x_{1}^{p_1}\hdots x_{n}^{p_n}f(x_{1},\hdots ,x_{n}) dx_{1}\,dx_{2}\hdots dx_{n}\\&=\int_{-\infty }^{\infty } \hdots  \int_{-\infty }^{\infty } x_{1}^{p_1}\hdots x_{n}^{p_n}f({\bf X}) d{\bf X}.
 \end{split}
 \end{equation}
 The moments of the density function $f({\bf X})$ exist if $f({\bf X})$ is piecewise continuous and has nonzero values only in a finite region of the space.  The moment $m_{p_1 \hdots  p_n}$ is called of order  $p=p_1+ \hdots + p_n$
 
 Similarly, the centered moments of order $p$ are defined by: 
 \begin{equation}
 \label{MomentsNDCentred}
 \mu_{p_1 \hdots  p_n}  =  \int_{-\infty }^{\infty } \hdots  \int_{-\infty }^{\infty } (x_{1}-\overline{x}_{1})^{p_1}\hdots (x_{n}-\overline{x}_{n})^{p_n}f({\bf X}) d{\bf X},
 \end{equation}
 where $(\overline{x}_{1}=\frac{m_{1,0 \hdots  0}}{m_{0,0 \hdots  0}},\,\overline{x}_{2}=\frac{m_{0,1,0\hdots  0}}{m_{0,0 \hdots  0}}, \hdots ,\overline{x}_{n}=\frac{m_{0 \hdots  0,n}}{m_{0,0 \hdots  0}})$ are the coordinates of the object gravity center. It is well known that the centered moments are invariant to translations in their respective n-dimensional space. In the sequel, the objects are considered centered at the frame origin, which means that   $\mu_{p_1 \hdots  p_n}= m_{p_1 \hdots  p_n}$. 
 Based on central moments, we propose an automatic scheme for deriving
 invariants to scale, to orthogonal transformations and finally to  affine  invariants. 
 The results presented in this paper for continuous case can be extended straightforwardly  moments of discrete distribution defined by
 \begin{equation}
 	 	\label{MomentsNDdiscrete} 	 
 	 		m_{p_1 \hdots p_n}  =  \sum  \hdots \sum x_{1}^{p_1}\hdots x_{n}^{p_n}f(x_{1},\hdots ,x_{n}) 
\end{equation}
 
 \section{Scale change}
 A scale change in  n-dimensional space is defined by the following transformation:
 \begin{equation}
 \label{Scalechange}
 {\bf X' }  = {\bf S}\,\,{\bf X}
 \end{equation}
 where ${\bf S}=diag(\sigma_1,\,\,\sigma_2,\hdots,\,\, \sigma_n)$. The scale change is called uniform if $\sigma_1=\sigma_2=\hdots=\sigma_n=\sigma$.  Invariants to uniform scale are easy to derive. Using (\ref{MomentsNDCentred})
 it is easy to show that  after a scale change defined by   $\sigma$, $\mu_{0,0 \hdots  0}$ and  moments of higher  
 $\mu_{p_1 \hdots  p_n}$ are  multiplied by  $\sigma^n$  and $\sigma^{n+p}$ respectively (remind $p=p_1+ \hdots + p_n$).  Therefore, the ratio $\frac{\mu_{p_1 \hdots  p_n}}{\mu_{0,0 \hdots  0}^{\frac{p+n}{n}}}$ in an invariant to uniform scale change.

 Let us now deal with  a non-uniform scale change. For that, let define the following  product of moments:
 \begin{equation}\label{MomentProduct}
 p_m=\prod_{i=1}^{l}\mu^{k_i}_{p^i_1 \hdots  p^i_n},
 \end{equation}
 such that 
 \begin{equation}
 \label{CondMomentProductaffine}
 \sum_{i=1}^{l} (p_j^i+1)k_i=d,\,\, \forall j=1,\hdots ,n
 \end{equation}
 and  where $d$ is a positive integer.  From (\ref{MomentProduct}), it can be seen that $p_m$ can be a product of $l$ moments of different orders.
 For each moment $\mu_{p^i_1 \hdots  p^i_n}$ of the product, the power on the coordinate $x_i$ is given by the integer $p_j^i$. This means that after the scale change (\ref{Scalechange}),   $x_i^{p_j^i}$ and  $x_i ^{p_j^i}dx_i$ are multiplied  by $\sigma_i^{p_j^i}$ and   $\sigma_i^{p_j^i+1}$ respectively. This results in a multiplication by $\prod_{j=1}^{n}\sigma_j^{ (p_j^i+1)}$  and $\prod_{j=1}^{n}\sigma_j^{ (p_j^i+1)k_i}$ on $\mu_{p^i_1 \hdots  p^i_n}$ and  $\mu^{k_i}_{p^i_1 \hdots  p^i_n}$ respectively. 
 Therefore,  after the scale change ${\bf S}$,  the whole product $p_m$ is multiplied by $\prod_{j=1}^{N}\sigma_j^{ \sum_{i=1}^{l} (p_j^i+1)k_i}=\prod_{j=1}^{n}\sigma_j^{ d} $ (if the condition (\ref{CondMomentProductaffine}) is satisfied).   Finally, since  after (\ref{Scalechange}), $\mu_{0,0 \hdots  0}$ is multiplied by  $\prod_{j=1}^{n}\sigma_j$, the ratio: 
 \begin{equation}
 \label{ScaleInvariantsGeneral}
 i_s=\frac{p_m}{{\mu_{0,0 \hdots  0}^{d}}}
 \end{equation}
 is an invariant to the scale change defined by  (\ref{Scalechange}).
 
 Let us now gives some example of invariants to scale  in 2D and 3D spaces:
 
 \begin{itemize}
 	\item 2D space: $\frac{\mu_{11}}{\mu_{00}^2},\,\,\frac{\mu_{20}\mu_{02}}{\mu_{00}^4},\,\,\frac{\mu_{30}\mu_{03}}{\mu_{00}^5},\,\,\frac{\mu_{21}\mu_{12}}{\mu_{00}^5},\,\,\frac{\mu_{22}}{\mu_{00}^3}$,\,\,$\frac{\mu_{40}\mu_{02}^2}{\mu_{00}^7},\,\,\frac{\mu^2_{30}\mu_{02}^3}{\mu_{00}^{11}}$
 	\item 3D space: $\frac{\mu_{111}}{\mu_{000}^2},\,\,\frac{\mu_{200}\mu_{020}\mu_{002}}{\mu_{000}^5},\,\, \frac{\mu_{200}\mu_{011}^2}{\mu_{000}^5},\,\,\frac{\mu_{020}\mu_{101}^2}{\mu_{000}^5},\,\,\frac{\mu_{002}\mu_{110}^2}{\mu_{000}^5} $
 		\item 4D space: $\frac{\mu_{1111}}{\mu_{0000}^2},\,\,\frac{\mu_{2000}\mu_{0200}\mu_{0020}\mu_{0002}}{\mu_{0000}^6},\,\, \frac{\mu_{2000}\mu_{0111}^3}{\mu_{0000}^6},\,\,\frac{\mu_{0020}\mu_{1101}^2}{\mu_{0000}^6} $
 \end{itemize}
In the next section, a new scheme  to obtain invariants to orthogonal transformation is given.
 \subsection{Invariants to orthogonal transformation}
  \subsection{Orthogonal transformation and rotation speed}
 An orthogonal transformation is defined by the relation:
 \begin{equation}
 \label{Rotations}
 {\bf X'}={\bf R}{\bf X}, 
 \end{equation}
 where  ${\bf R}$ is a rotation matrix that have to satisfy ${\bf R}{\bf R}^\top={\bf I}$  and $det({\bf R})=1$ (${\bf I}$ is the identity matrix in the n-dimensional space). One can show that a moment after a rotation can be expressed as linear 
 combination of moments of the same order, where the coefficients of those combinations  are  polynomials    on rotation matrix entries.  The rotation invariance  of some functions of moments are obtained thanks to  the orthogonality condition of  ${\bf R}$.
 The latter is composed of constraints that are nothing but polynomials of orders $2$ on the rotation matrix  entries. For this reason, 
 it is not possible to build invariants to rotations from moments of odd orders without putting them to some even  power.

 Rather basing our reasoning on  the rotation matrix and its orthogonality condition, it is possible  to build invariants based on rotation speed (Tensor) in very simple way. For that, let us consider that an n-dimensional solid  undergoes a transformation in 
 time. If a rotational speed is applied, the speed of each point  of an n-dimensional object is given by:
  \begin{equation}
\label{Rotationsvar}
{\dot{\bf X}}={\bf L}{\bf X}
\end{equation}
where ${\bf L}$ is an antisymmetric matrix  defined by rotational speeds in the 2D planes built  by two different axes. 
For instance ${\bf L}$ is defined as follows:
\begin{itemize}
	\item In 2D space:
	\begin{equation}\label{A2D}
	{\bf L_{2D}}=
	\left[
		\begin{array}{cc}
	0&-\omega\\
	\omega &0
	\end{array}
	\right]
	\end{equation}
	where $\omega$ is the rotation speed in the 2D plane.
		\item In 3D space:
	\begin{equation}\label{A3D}
	{\bf L_{3D}}=
	\left[
	\begin{array}{ccc}
	0&-\omega_3&\omega_2\\
	\omega_3 &0&-\omega_1\\
	-\omega_2&\omega_1&0
	\end{array}
	\right]
	\end{equation}
	where $\omega_1$,  $\omega_2$,  $\omega_3$ are respectively the rotation speeds in the planes
	$yz$, $zx$ and $xy$. The scalars $\omega_1$,  $\omega_2$,  $\omega_3$ can be also be  called respectively  the rotation speeds around
	the x-axis, the y-axis and the z-axis in the case of 3D space. 
		\item In 4D space, let us consider that the 4D frame has $4-axes $ $x$, $y$, $z$ and $w$.  The antisymmetric matrix $\bf A_{4D} $ can be defined by:
	\begin{equation}\label{A4D}
	{\bf L_{4D}}=
	\left[
	\begin{array}{cccc}
0& \omega_1&-\omega_2&\omega_3\\
-\omega_1&0&\omega_4&-\omega_5\\
\omega_2&-\omega_4&0&\omega_6\\
-\omega_3&\omega_5&-\omega_6&0
	\end{array}
	\right]
	\end{equation}
	where $\omega_1$,  $\omega_2$,  $\omega_3$, $\omega_4$,  $\omega_5$,  $\omega_6$ are respectively the rotation speeds in the planes	$xy$, $xz$, $xw$, $yz$, $yw$ and $zw$.	
\end{itemize}
Actually, every  two  frame axis of an n-dimensional space specifies a   2D plane. Therefore, there exist $\frac{n!}{ (n-2)!2!}$  different planes and then the same number of rotational speeds.  

  \subsection{Moment time variation and rotational speeds}
 After taking the derivative of  (\ref{MomentsND}), the time variation of a moment $ m_{p_1 \hdots  p_n}$ is obtained by \cite{Tahri03b}, \cite{tahri2004utilisation}:
 \begin{equation}
 \label{dmp1ppn}
 \begin{split}
  \dot{m}_{p_1 \hdots  p_n}&=\sum_{i=0}^{n}\int_{-\infty }^{\infty } \hdots  \int_{-\infty }^{\infty }p_i\dot{x}_i x^{p_i-1}\prod_{j=1, j\neq i}^{n} x_{1}^{p_1}\hdots x_{n}^{p_n}f(x_{1},\hdots ,x_{n}) dx_{1}\,dx_{2}\hdots dx_{n}\\&+\int_{-\infty }^{\infty } \hdots  \int_{-\infty }^{\infty } x_{1}^{p_1}\hdots x_{n}^{p_n}\dot{f}(x_{1},\hdots ,x_{n}) dx_{1}\,dx_{2}\hdots dx_{n}\\
  &+=\int_{-\infty }^{\infty } \hdots  \int_{-\infty }^{\infty } x_{1}^{p_1}\hdots x_{n}^{p_n}f(x_{1},\hdots ,x_{n})(\sum_{i=0}^{n}
  \frac{\partial \dot{x}_i}{\partial x_i}) dx_{1}\,dx_{2}\hdots dx_{n}
  \end{split}
 \end{equation}
Since the matrix ${\bf L}$ is antisymmetric (the diagonal entries are null), we have $ \frac{\partial \dot{x}_i}{\partial x_i}=0$.
This implies that the third term of  (\ref{dmp1ppn}) vanishes when rotational speeds are applied to the object.  The second term of 
(\ref{dmp1ppn}) vanishes as well if we assume that the time derivative of density function $\dot{f}(x_{1},\hdots ,x_{n})=0$.  The latter assumption has been also made in \cite{reiss1991revised} and \cite{mamistvalov1998n} to prove the the Revised FTMI. Assuming $\dot{f}(x_{1},\hdots ,x_{n})=0$ means that applying a rotation on the object point does not change its corresponding  density function value.  According to (\ref{Rotationsvar}), we have  $\dot{x}_i=\sum_{j=0}^{n}l_{ij}x_j$ ($l_{ij}$ are the entries of the matrix ${\bf L}$).  Combining this with  (\ref{dmp1ppn}), the time variation of moment caused by 
 rotational speeds can be obtained by:

 \begin{equation}
\label{dmp1ppnsimp}
\dot{m}_{p_1 \hdots  p_n}=\sum_{i=0}^{n}\sum_{j=0}^{n}p_il_{ij}{m}_{p_1,\hdots ,p_i-1,\hdots ,p_j+1, p_n}. 
\end{equation}
Let us now consider for instance moments  from 2D, 3D and 4D spaces. 
\begin{itemize}
	\item In 2D space, using  (\ref{A2D}) leads to:
	\begin{equation}
	\label{dmp1ppnsimp2D}
	\dot{m}_{p_1p_2}=(-p_1m_{p_1-1,p_2+1}+p_2m_{p_1+1,p_2-1})\,\,\omega.
	\end{equation}
		\item In 3D space, using  (\ref{A3D}) leads to:
	\begin{equation}
	\label{dmp1ppnsimp3D}
	 \begin{split}
	\dot{m}_{p_1p_2p_3}&=(p_2m_{p_1,p_2-1,p_3+1}-p_3m_{p_1,p_2+1,p_3-1})\,\,\omega_1\\
	             &+(p_3m_{p_1-1,p_2,p_3-1}-p_1m_{p_1-4,p_2,p_3+1})\,\,\omega_2\\
	             &+(p_1m_{p_1-1,p_2+1,p_3}-p_2m_{p_1+1,p_2-1,p_3})\,\,\omega_3.
	 \end{split}
	\end{equation}
		\item In 4D space, using  (\ref{A4D}) leads to:
			\begin{equation}
			\label{dmp1ppnsimp4D}
			\begin{split}
			\dot{m}_{p_1p_2p_3p_4}&=(p_1m_{p_1-1,p_2+1,p_3,p_4}-p_2m_{p_1+1,p_2-1,p_3,p_4})\,\,\omega_1\\&+(p_3m_{p_1+1,p_2,p_3-1,p_4}-p_1m_{p_1-1,p_2,p_3+1,p_4})\,\,\omega_2\\
		    &+(p_1m_{p_1-1,p_2,p_3,p_4+1}-p_4m_{p_1+1,p_2,p_3,p_4-1})\,\,\omega_3\\&+(p_2m_{p_1,p_2-1,p_3+1,p_4}-p_3m_{p_1,p_2+1,p_3-1,p_4})\,\,\omega_4\\
		      &+(p_4m_{p_1,p_2+1,p_3,p_4-1}-p_2m_{p_1,p_2-1,p_3,p_4+1})\,\,\omega_5\\&+(p_3m_{p_1,p_2,p_3-1,p_4+1}-p_4m_{p_1,p_2,p_3+1,p_4-1})\,\,\omega_6.
		    			\end{split}
	\end{equation}
\end{itemize}

In the next section, a simple method to determine invariant to rotations using the relation  between the time variation of moments 
and the rotational speeds.  

\subsection{Time variation of Moment vectors}
Let  ${\bf v}^1_p$ be  the  vector composed by all moments of order $p$. Let ${\bf v}^k_p$ be the monomials  vector of degree  $k$ computed from ${\bf v}^1_p$. As an example from 2D space for $p=2$ and $k=2$:   $${\bf v}^1_2=[m_{20},\,\,m_{11},\,\,m_{02}]^\top,\,\,{\bf v}^2_2=[m_{20}^2,\,\,m_{20}m_{11},\,\,m_{20}m_{02},\,\,m^2_{11},\,\,m_{02}m_{11},\,\,m_{02}^2]^\top$$ Let us also define moment vectors mixing moments of different orders  ${\bf v}^k_p$ and  ${\bf v}^{k'}_{p'}$. For instance,  by multiplying the entries of  ${\bf v}^1_2$ and ${\bf v}^2_3$, a new moment vector can be obtained  as
\[
\bf{m}^{(1,2)}_{(2,3)}=
\left[
\begin{array}{ccccc}
m_{20}m^2_{30}&m_{20}m_{30}m_{21}&m_{20}m_{30}m_{12}&\hdots&m_{02}m^2_{03}.
\end{array}
\right]^\top.
\]
Moment vectors from three different orders or more can also be defined. For instance $\bf{m}^{(k,k',k'')}_{(p,p',p'')}$ by all possible products between the entries  of the vectors  $\bf{m}^{k}_{p}$, $\bf{m}^{k'}_{p'}$ and $\bf{m}^{k''}_{p''}$.
	
Using (\ref{dmp1ppnsimp}), it can be concluded that the time variation of each entry  of ${\bf v}^1_p$ is a linear combination 
of the other entries and the rotational speed. This implies that the time variation of the moment vector ${\bf v}^1_p$ can be written as follow:
\begin{equation}
\label{dmp1dt}
\dot{\bf v}^1_p=\sum_{i=0}^{\frac{n!}{ (n-2)!2!}}{\bf L}^{\omega_i}_{{\bf v}^1_p}\,{\bf v}^1_p\,\omega_i
\end{equation}
where ${\bf L}^{\omega_i}_{{\bf v}^1_p}$ are  matrices of integers. 
As a  simple example from 2D space, let  us consider  ${\bf v}^1_2=[m_{20},\,\,m_{11},\,\,m_{02}]^\top$. For the latter vector, using 
 (\ref{dmp1ppnsimp2D}), we obtain:
 \begin{equation}
\label{dm21dt}
\dot{\bf v}^1_2={\bf L}^{\omega}_{{\bf v}^1_2}{\bf v}^1_2=\left[
\begin{array}{ccc}
0&-2&0\\
1&0&-1\\
0&2&0
\end{array}
\right]{\bf v}^1_2\,\,\omega.
\end{equation}

Since $\bf{m}^{k}_{p}$ is defined by the monomials on $\bf{m}^{1}_{p}$, the time derivative $\dot{\bf{m}}^{k}_{p}$ can be expressed 
as linear combination of the  $\bf{m}^{k}_{p}$ entries. We have then 
\begin{equation}
\label{dmpkdt}
\dot{\bf v}^k_p=\sum_{i=0}^{\frac{n!}{ (n-2)!2!}}{\bf L}^{\omega_i}_{{\bf v}^k_p}\,{\bf v}^k_p\,\omega_i.
\end{equation}
Since ${\bf L}^{\omega_i}_{{\bf v}^1_p}$ are  matrices  of integers, ${\bf L}^{\omega_i}_{{\bf v}^k_p}$ are
 also of the same nature.   This is also true for any  vectors  ${\bf{v}}^{(k,k',k'')}_{(p,p',p'')}$  mixing moments of different orders. In the following, it will be shown that it is easy to build invariants to rotations as linear combination 
of the entries of  the previous kinds of moment vectors. 
\subsubsection{Rotation invariants}
Let us consider  a scalar $i_{r^{k}_p}=\boldsymbol{\alpha}^{k\top}_p{\bf v}^k_p$, where $\boldsymbol{\alpha}^{k}_p$ is a vector of coefficients.
Therefore, the time derivative of $i_{r^{k}_p}$ is defined by:
\begin{equation}
\label{dindt}
\frac{di_{r^{k}_p}}{dt}=\boldsymbol{\alpha}^{k\top}_p \dot{\bf v}^k_p.
\end{equation}
 Combining (\ref{dindt}) with (\ref{dmpkdt}) gives:
 \begin{equation}
 \label{dindt2}
 \frac{di_{r^{k}_p}}{dt}=\sum_{i=0}^{\frac{n!}{ (n-2)!2!}}\boldsymbol{\alpha}^{k\top}_p{\bf L}^{\omega_i}_{{\bf v}^k_p}\,{\bf v}^k_p\,\omega_i.
 \end{equation}
The scalar $i_{r^{k}_p}$ is invariant rotation if:
\[
\boldsymbol{\alpha}^{k\top}_p{\bf L}^{\omega_i}_{{\bf v}^k_p}\,{\bf v}^k_p=0, \,\,\, \forall i=1,\hdots,{\frac{n!}{ (n-2)!2!}}.
\]
This implies the following condition:
 \begin{equation}
\label{conrotinvariance}
{\bf L}^{\omega_i\,\,\top}_{{\bf v}^{k}_p}\boldsymbol{\alpha}^{k}_p={\bf{0}}, \,\,\,\forall i=1,\hdots,{\frac{n!}{ (n-2)!2!}}
\end{equation}
If we take again the simple moment vector  ${\bf v}^1_2=[m_{20},\,\,m_{11},\,\,m_{02}]^\top$, we have:
\[
{\bf L}^{\omega\top}_{{\bf v}^1_2}=\left[
\begin{array}{ccc}
0&1&0\\
-2&0&2\\
0&-1&0
\end{array}
\right]
\]
The null space of ${\bf L}^{\omega\top}_{{\bf v}^1_2}$ is:
\[
\boldsymbol{\alpha}^1_2=[1\,\,\,\, 0\,\,\,\, 1]^\top
\]
Which gives as invariant $i_{r^1_2}=m_{20}+m_{02}$.

 In the next section, the  results about  both invariant to rotations and   those to non-uniform scale change will be used to derive affine invariants in a very straightforward manner. 

\section{Affine invariants}
\subsection{Proposed method}
Since rotation transformation is a subgroup of affine ones, one can conclude easily that  an  affine invariant
is an  invariant to rotation. Let us consider that the affine transformation is defined by a matrix ${\bf A}$, such that:
\begin{equation}
\label{affinetrans}
{\bf X'}= {\bf A}{\bf X}.
\end{equation} 
Here, the translation has  not been considered since we consider that the object is centered.  Actually, using singular value decomposition, any non-singular   matrix ${\bf A}$ can be decomposed as:
\[
{\bf A}={\bf R}_2\,\, {\bf S}\,\,{\bf R}_1, 
\]
where  ${\bf R}_1$ and ${\bf R}_2$ are rotations and  ${\bf S}$ is non uniform scale change. This means that an affine transformation is equivalent to  a rotation followed by a non-uniform scale change then by another  rotation.
To explain the proposed approach, let us consider building affine invariants from: $${\bf v}^2_2=[m_{20}^2,\,\,m_{20}m_{11},\,\,m_{20}m_{02},\,\,m^2_{11},\,\,m_{02}m_{11},\,\,m_{02}^2]^\top$$
 In the vector 
of ${\bf v}^2_2$, two  entries hold the condition to be invariant to non-uniform scale change  (\ref{CondMomentProductaffine}), which are  $m^2_{11}$ and $m_{20}m_{02}$. The vector ${\bf v}_{s^2_2}=\frac{1}{m^4_{00}}[m_{20}m_{02}\,\,\, m^2_{11}]^\top$  composed 
by these moments products is an invariant to non-uniform scale change.  Using  (\ref{dmp1ppnsimp2D}), their time derivatives are given  by:
\begin{equation}
\label{affineexample1}
\left\{
\begin{split}
&\frac{d(m_{20}m_{02})}{dt}=\dot{m}_{20}{m}_{02}+{m}_{20}\dot{m}_{02}=(-2{m}_{02}m_{11}+2{m}_{20}m_{11})\omega\\
&\frac{d(m^2_{11})}{dt}=2m_{11}\dot{m}_{11}=(2m_{11}m_{20}-2m_{11}m_{02})\omega
\end{split}
\right.
\end{equation}
Since $m^4_{00}$ is  invariant to rotations, from (\ref{affineexample1}) we obtain:
\begin{equation}
\label{affineexample1_2}
\dot{{\bf v}}_{s^2_2}=\frac{1}{m^4_{00}}
\left[
\begin{array}{c}
\frac{d(m_{20}m_{02})}{dt}\\
\frac{d(m^2_{11})}{dt}
\end{array}
\right]={\bf L_{{\bf v}_{s^2_2}}} \frac{{\bf v}^2_2}{m^4_{00}}  \,\,\omega=\left[
\begin{array}{cccccc}
0&2&0&0&-2&0\\
0&2&0&0&-2&0
\end{array}
\right]\frac{{\bf v}^2_2}{m^4_{00}} \,\, \omega
\end{equation}
Let us consider $i_{a^2_2}=\boldsymbol{\alpha}^{2\top}_2 {\bf v}_{s^2_2}$.
where  $\boldsymbol{\alpha}^{2\top}_2$ is vector of coefficients of the same size as
 ${\bf v}_{s^2_2}$. As it has been shown in  the previous section, $i_{a^2_2}$ is rotation invariant if $\boldsymbol{\alpha}^{2\top}_2 {\bf L_{{\bf v}_{s^2_2}}}={\bf 0}$. Therefore, it can be obtained that 
$\boldsymbol{\alpha}=ker({\bf L^\top_{{\bf v}_{s^2_2}}})=[1 -1]^\top$, which leads to the  invariant:
\begin{equation}
\label{affineexample1_3}
i_{a^2_2}=\frac{m_{20}m_{02}-m^2_{11}}{m^4_{00}}
\end{equation}
which is the well known and the simplest affine invariant in 2D space. 
Actually, this scheme can be applied to any moment vector in a n-dimensional space
as follow:
\begin{itemize}
	\item Consider a moment vector ${\bf v}^k_p$ (or one combining moments of different orders ${\bf{v}}^{(k,k',k'')}_{(p;p';p'')}$),
	\item Build a new moment vector ${\bf v}_s$ by selecting only entries of   ${\bf v}^k_p$ if divided by the required 
	power of $m_{0\hdots0}^d$ become invariants to non-uniform scale change.
	\item The time derivative of  ${\bf v}_s$ when rotational speeds are applied to the object can be written as:
	\begin{equation}
	\label{dvspkdt}
	\dot{\bf v}_s=\frac{1}{m_{0\hdots0}^d}\sum_{i=0}^{\frac{n!}{ (n-2)!2!}}{\bf L}^{\omega_i}_{{\bf v}_s}\,{\bf v}^k_p\,\omega_i.
	\end{equation}
	 \item Compute  $i_a=\boldsymbol{\alpha}^\top {\bf v}_s$ where: 
	 \[
	 \boldsymbol{\alpha}^\top{\bf L}^{\omega_i}_{{\bf v}_s}\,{\bf v}^k_p=0, \,\,\, \forall i=1,\hdots,{\frac{n!}{ (n-2)!2!}}.
	 \]
\end{itemize} 
Since each entry of  ${\bf v}_s$ is an invariant to non-uniform scale change and  the linear combination $i_a=\boldsymbol{\alpha}^\top {\bf v}_s$ is invariant to rotation, therefore $i_a$ is an affine invariant.
In the following,  some example invariant to rotations and to affine transformations in 4D space. 
 
\begin{small} 
\begin{equation}
\label{ir32d4}
\begin{split}
&i_{r^3_2}(1)=m_{0012}^2- m_{0012}m_{0210 }- m_{0012}m_{2010 }- m_{0030}m_{0012 }+ m_{0021}^2- m_{0021}m_{0201 }\\
&- m_{0021}m_{2001 }- m_{0003}m_{0021 }+ m_{0102}^2- m_{0102}m_{0120 }- m_{0102}m_{2100 }- m_{0300}m_{0102 }\\
&+3m_{0111}^2+ m_{0120}^2- m_{0120}m_{2100 }- m_{0300}m_{0120 }+ m_{0201}^2- m_{0201}m_{2001 }- m_{0003}m_{0201 }\\
&+ m_{0210}^2- m_{0210}m_{2010 }- m_{0030}m_{0210 }+ m_{1002}^2- m_{1002}m_{1020 }- m_{1002}m_{1200 }\\
&- m_{3000}m_{1002 }+3m_{1011}^2+ m_{1020}^2- m_{1020}m_{1200 }-m_{3000}m_{1020}+3m_{1101}^2+3m_{1110}^2\\
&+ m_{1200}^2- m_{3000}m_{1200 }+ m_{2001}^2- m_{0003}m_{2001 }+ m_{2010}^2- m_{0030}m_{2010 }+ m_{2100}^2\\
&- m_{0300}m_{2100}\\
&i_{r^3_2}(2)=3m_{0003}m_{0021 }+3m_{0012}m_{0030 }+3m_{0003}m_{0201 }+3m_{0012}m_{0210 }+3m_{0021}m_{0201 }\\
&+3m_{0102}m_{0120 }+3m_{0030}m_{0210 }+3m_{0102}m_{0300 }+3m_{0120}m_{0300 }+3m_{0003}m_{2001 }+3m_{0012}m_{2010 }\\
&+3m_{0021}m_{2001 }+3m_{1002}m_{1020 }+3m_{0030}m_{2010 }+3m_{0102}m_{2100 }+3m_{0201}m_{2001 }+3m_{1002}m_{1200 }\\
&+3m_{0120}m_{2100 }+3m_{0210}m_{2010 }+3m_{1020}m_{1200 }+3m_{0300}m_{2100 }+3m_{1002}m_{3000 }+3m_{1020}m_{3000 }\\
&+3m_{1200}m_{3000 }+ m_{0003}^2+ m_{0030}^2-3m_{0111}^2+ m_{0300}^2-3m_{1011}^2-3m_{1101}^2-3m_{1110}^2+ m_{3000}^2
\end{split}
\end{equation}
\end{small} 

\begin{equation}
\label{ia4d24}
\begin{split}
&i_{a^4_2}=\frac{1}{m^6_{0000}}(m_{0011}^2m_{1100}^2 - m_{0200}m_{2000}m_{0011}^2 + 2m_{2000}m_{0011}m_{0101}m_{0110 }\\ &-2m_{0011}m_{0101}m_{1010}m_{1100 }-2m_{0011}m_{0110}m_{1001}m_{1100 } +2m_{0200}m_{0011}m_{1001}m_{1010 }\\
&+ m_{0101}^2m_{1010}^2 - m_{0020}m_{2000}m_{0101}^2 - 2m_{0101}m_{0110}m_{1001}m_{1010 } + m_{0110}^2m_{1001}^2\\
& +2m_{0020}m_{0101}m_{1001}m_{1100 } + 2m_{0002}m_{0110}m_{1010}m_{1100 }+m_{0002}m_{0020}m_{0200}m_{2000} \\
&- m_{0020}m_{0200}m_{1001}^2 - m_{0002}m_{0200}m_{1010}^2- m_{0002}m_{2000}m_{0110}^2  - m_{0002}m_{0020}m_{1100}^2)
\end{split}
\end{equation}

\subsection{Discussion}
The set of constraints given by (\ref{conrotinvariance}) means that if $i_r$ is  invariant to rotation, its time variation 
caused by any of the ${\frac{n!}{ (n-2)!2!}}$ rotational speeds is null. Let us consider the case of 3D space. 
In that case, as it has been mentioned above, there exist  $\omega_1$,  $\omega_2$,  $\omega_3$ are respectively the rotation speeds in the planes $yz$, $zx$ and $xy$ respectively. Let us consider a vector of 3D moments ${\bf v}^k_p$ and   $i_{r_p^k}=\boldsymbol{\alpha}^{k\top}_p {\bf v}^k_p$ an invariant to rotations. According to $(\ref{conrotinvariance})$, $\boldsymbol{\alpha}$ satisfies the three conditions:
\[
\begin{split}
&(c_1)\,\,\, {\bf L}^{\omega_1\top}_{{\bf v}^k_p}\,\,\boldsymbol{\alpha}^k_p={\bf 0}\\
&(c_2)\,\,\,{\bf L}^{\omega_2\top}_{{\bf v}^k_p}\,\,\boldsymbol{\alpha}^k_p={\bf 0}\\
&(c_3)\,\,\,{\bf L}^{\omega_3\top}_{{\bf v}^k_p}\,\,\boldsymbol{\alpha}^k_p={\bf 0}.
\end{split}
\] 
Each of the three constraints implies the invariance of $i_r$ with respect to the rotations in the plane corresponding to $\omega_i$. This means  that the constraints $c_1$ and $c_2$ implies invariance with respect to rotations in the planes 
$yz$ and $xz$. Actually, if $c_1$ and $c_2$ are satisfied, there is no need to check $c_3$ since a rotation in the $xy$ plane 
can be expressed as two consecutive rotations  in the plane $xz$ to bring the x-axis to the z-axis position then a rotation 
in the $yz$ plane. In n-dimensional space as well, the vector of coefficients  $\boldsymbol{\alpha}$ has to  ensure 
invariance with respect to rotations only in  $n-1$ well chosen planes instead of ${\frac{n!}{ (n-2)!2!}}$.  To show that,
let us consider that n-dimensional has n axis named by $n$ Latin letter $[x\,\,y\,\,z\,\,w,\,\,\hdots]$.
The x-axis combined to each of the $n-1$ other ones form  $n-1$ planes ($xy,xz,xw,....$) and offer $n-1$  possible planar rotations.
Actually, any other rotation can be obtained as applying two consecutive rotations  from   $n-1$  considered above. For instance, a rotation 
in the  in the $zw$ plane can be obtained as two consecutive rotations  in the plane $xz$ followed  another in $xw$ plane.

In practice, the matrices  ${\bf L}^{\omega_i\top}_{{\bf v}^k_p}$ are very sparse. First, when they are calculated and stored, one can consider only non-null values. Second, the vector of coefficients can be computed using  null space adapted 
to sparse matrix. This becomes  necessary for large moment vectors. To give an idea how sparse are those matrix, let us consider the 3D moment vector  ${{\bf v}^4_3}$, which is of size equal to $715$.    Let us consider computing an affine invariant from this vector.
In that case, the entries of  ${{\bf v}^4_3}$ satisfying  the condition to be invariant to non-uniform scale change  (\ref{CondMomentProductaffine}) are  only $25$.  From these entries, an invariant vector  to non-uniform scale can be defined by:
\[
 \begin{split}
&{\bf v}_{s^4_3}\!=\![m_{003}m_{021}m_{120}m_{300},\,
m_{012}^2m_{120}m_{300},\,
m_{003}m_{030}m_{111}m_{300},
m_{012}m_{021}m_{111}m_{300}\\&
m_{012}m_{030}m_{102}m_{300},
m_{021}^2m_{102}m_{300},\,
m_{003}m_{021}m_{210}^2,\,
m_{003}m_{030}m_{201}m_{210},\,
m_{012}^2m_{210}^2,\,\\&
m_{012}m_{021}m_{201}m_{210},\,\,\,\,\,
m_{003}m_{111}m_{120}m_{210},\,\,\,\,
m_{012}m_{102}m_{120}m_{210},\,
m_{012}m_{111}^2m_{210},\\&
m_{021}m_{102}m_{111}m_{210},\,
m_{030}m_{102}^2m_{210},\,\,
m_{012}m_{030}m_{201}^2,\,
	m_{021}^2m_{201}^2,\,\,
	m_{003}m_{120}^2m_{201},\,\\&
	m_{012}m_{111}m_{120}m_{201},\,\,\,\,
	m_{021}m_{102}m_{120}m_{201},\,\,\,\,
	m_{030}m_{102}m_{111}m_{201},\,\,\,\,
	m_{021}m_{111}^2m_{201},\\&
	m_{102}^2m_{120}^2,\,\,
	m_{102}m_{111}^2m_{120},\,\,
	m_{111}^4	]^\top\frac{1}{m^8_{000}}
	\end{split}
\]
The time derivative of ${\bf v}_{s^4_3}$ can be written under the form:
\begin{equation}
\label{dvspkdtexample2}
\dot{\bf v}_{s^4_3}=\frac{1}{m_{000}^8}\left({\bf L}^{\omega_1}_{{\bf v}_s}\,{\bf v}^4_3\,\omega_1 + {\bf L}^{\omega_2}_{{\bf v}_s}\,{\bf v}^4_3\,\omega_2+ {\bf L}^{\omega_3}_{{\bf v}_s}\,{\bf v}^4_3\,\omega_3   \right),
\end{equation}
where ${\bf L}^{\omega_i}_{{\bf v}_s}$ are matrices of integers of size $25\times 715$. If we consider obtaining from ${\bf v}_{s^4_3}$ 
an affine invariant under the form of $i_a=\boldsymbol{\alpha}^\top {\bf v}_s$, the coefficient vector  will be obtained by:
\[\boldsymbol{\alpha}=ker({\bf L}^{\omega_1,\omega_2}_{{\bf v}_s}),\,\, \mbox{where:\,\,} {\bf L}^{\omega_1,\omega_2}_{{\bf v}_s}=\left[\begin{array}{c}
{\bf L}^{{\omega_1}^\top}_{{\bf v}_s}\\
{\bf L}^{{\omega_2}^\top}_{{\bf v}_s}
\end{array}
\right]
\]
The matrix ${\bf L}^{\omega_1,\omega_2}_{{\bf v}_s}$ is of size $1430\times 25$, but it is very sparse. Actually, after removing all the rows of  ${\bf L}^{\omega_1,\omega_2}_{{\bf v}_s}$ that contain only null values,   the vector  of coefficients  can  be obtained as the null space of a matrix of reduced size equal to  $84\times 25$. Besides, even  after this consequent size reduction, the resulting reduced size matrix is still  very sparse. The obtained formula of $i_a$ is given  by:
\begin{small}
\[
\begin{split}
&i_{a_3^4}=\frac{1}{m^8_{000}}(\!-\!m_{300}m_{012}^2m_{120 }\!+\! m_{012}^2m_{210}^2\!+\! m_{300}m_{012}m_{021}m_{111 }- m_{012}m_{021}m_{201}m_{210 }\\&\!-\! m_{012}m_{102}m_{120}m_{210 }\!+\! m_{030}m_{300}m_{012}m_{102 }- 2m_{012}m_{111}^2m_{210 }\!+\! 3m_{012}m_{111}m_{120}m_{201 }\\&- 
m_{030}m_{012}m_{201}^2\!-\! m_{300}m_{021}^2m_{102 }\!+\! m_{021}^2m_{201}^2\!+\! 3m_{021}m_{102}m_{111}m_{210 }\!-\! 2m_{021}m_{111}^2m_{201 }\\&\!-\! m_{021}m_{102}m_{120}m_{201 }\!+\! m_{003}m_{300}m_{021}m_{120 }\!-\! m_{003}m_{021}m_{210}^2\!+\! m_{102}^2m_{120}^2\!-\! m_{030}m_{102}^2m_{210 }
\\&\!-\! 2m_{102}m_{111}^2m_{120 }\!+\! m_{030}m_{102}m_{111}m_{201 }\!+\! m_{003}m_{111}m_{120}m_{210 }\!-\! m_{003}m_{030}m_{300}m_{111 }\\&\!-\! m_{003}m_{120}^2m_{201 }\!+\! m_{003}m_{030}m_{201}m_{210}\!+\! m_{111}^4)
\end{split}
\]
\end{small}
From the formula of the previous affine invariant, it can be noticed 
 the coefficients of many entries of ${\bf v}_{s^4_3}$ have the sames values. 
 Actually, the entries  of ${\bf v}_{s^4_3}$  can be divided to 10 groups:

\begin{itemize}
	\item Group 1:  $m_{300}m_{012}^2m_{120 }$, $m_{300}m_{021}^2m_{102 }$, $m_{003}m_{120}^2m_{201 }$, $m_{030}m_{012}m_{201}^2$, $ m_{003}m_{021}m_{210}^2$, $m_{030}m_{102}^2m_{210 }$;
	\item Group 2: $ m_{021}^2m_{201}^2$, $m_{012}^2m_{210}^2$, $m_{102}^2m_{120}^2$;
	\item group 3:  $m_{030}m_{102}m_{111}m_{201 }$,  $m_{003}m_{111}m_{120}m_{210 }$, $ m_{300}m_{012}m_{021}m_{111 }$;
	\item Group 4: $m_{012}m_{111}^2m_{210 }$, $m_{021}m_{111}^2m_{201 }$, $m_{102}m_{111}^2m_{120 }$;
	\item Group 5: $m_{003}m_{300}m_{021}m_{120 }$, $m_{003}m_{300}m_{021}m_{120 }$, $m_{030}m_{300}m_{012}m_{102 }$;
	\item Group 6: $m_{012}m_{021}m_{201}m_{210 }$, $m_{012}m_{102}m_{120}m_{210 }$;
	\item Group 7: $m_{021}m_{102}m_{111}m_{210 }$, $m_{012}m_{111}m_{120}m_{201 }$;
	\item Group 8: $m_{012}m_{111}m_{120}m_{201 }$;
	\item Group 9:  $ m_{111}^4$;
  	\item Group 10: $m_{003}m_{030}m_{300}m_{111 }$;
\end{itemize}
Any member of a group can be obtained by coordinates  switching   from another of the same group. Since $i_{a_3^4}$ is  an invariant to rotations, the entries of the same group will  have the same coefficient.  This reduce the number of unknown to $10$ coefficients 
instead of $25$. 
\section{Redundancy between invariants}
As it has been shown previously, invariants to rotations or to affine transformation can be obtained 
as linear combination of the entries of moment vectors.  The vectors  of coefficients $\boldsymbol{\alpha}$ defining 
the linear combinations are obtained as the null space of a matrix of integers.
Therefore,  by construction, the set of invariants that can be obtained from a vector of moment are linearly independent. 
However, as dealt with  in  \cite{suk2011affine} for instance,   dependency between invariants of higher order and of lower should
 be removed.
 
 Actually, the form and the proposed method  used to obtain invariants simplify  redundancy elimination. Let us consider an example
 from 2D space where rotation invariants are to be computed from the moment vectors ${\bf v}^1_2$, ${\bf v}^2_5$ and ${\bf v}^{1,2}_{2,5}$. The products of  two  invariants computed respectively from ${\bf v}^1_2$ and from  ${\bf v}^2_5$ are invariants of the same kind as  those computed from  ${\bf v}^{(1,2)}_{(2,5)}$.  Therefore, while computing invariant from the latter 
 moment vector,  it is necessary to eliminate those obtained from the two first. The elimination of such dependency can be achieved as follow:
\begin{itemize}
	\item  Compute the vector of coefficients $\boldsymbol{\alpha}^1_2$ from ${\bf v}^1_2$. We obtain 
	\begin{equation}
	\label{in21} 
	i_{r^1_2}=m_{20}+m_{02}	
	\end{equation}
	\item  Compute the vector of coefficients $\boldsymbol{\alpha}^2_5$ from ${\bf v}^2_5$. We obtain 3 invariants:
	\begin{small}
	\begin{equation}
	\label{in52}
	\hspace{-.5cm}
	\begin{split}
	i_{r^2_5(1)}&= 3m_{23}^2 - 4m_{41}m_{23} \!+\! 3m_{32}^2 - 4m_{14}m_{32} \!+\! m_{05}m_{41} \!+\! m_{14}m_{50}\\
	i_{r^2_5(2)}&\!=\! m_{14}m_{32} - m_{05}m_{41} - m_{05}m_{23} - m_{14}m_{50} \!+\! m_{23}m_{41} - m_{32}m_{50} \!+\! m_{14}^2 \!+\! m_{41}^2\\
	i_{r^2_5(3)}&=15m_{05}m_{23} \!+\! 5m_{05}m_{41} \!+\! 25m_{14}m_{32} \!+\! 5m_{14}m_{50} \!+\! 25m_{23}m_{41} + 15m_{32}m_{50}  \\&\!+\! 3m_{05}^2\!+\! 3m_{50}^2\\
	\end{split}
	\end{equation}
\end{small}
	\item  The product of  invariants given by (\ref{in21}) and  (\ref{in52}) leads to  3 rotation invariants of the same kind as those that can be  obtained 
	from ${\bf v}^{1,2}_{2,5}$. They can be written under the form $\boldsymbol{\beta}^\top {\bf v}^{(1,2)}_{(2,5)}$.  Without taking into account this dependency, the null space of
	 ${\bf L}^\top_{{\bf v}^{(1,2)}_{(2,5)}}$ allows to obtain  $9$ invariants to rotations $i_{r^{(1,2)}_{(2,5)}}$  from ${\bf v}^{(1,2)}_{(2,5)}$. The dependency can be easily removed by computing the null space of:
\[
\boldsymbol{\alpha}=ker\left(
\left[
\begin{array}{c}
\boldsymbol{\beta}^\top\\
{\bf L}^\top_{{\bf v}^{1,2}_{2,5}}
\end{array}
\right]
\right)
\] 
which gives  only $6$ invariants  independent of $i_{r^1_2}$ and $i_{r^2_5}$.  	From those $6$
	invariants, only $1$ affine invariant can be derived:
	\[\hspace{-.6cm}
		\begin{split}
	i_{a^{(1,2)}_{(2,5})}\!=\,&\frac{1}{m^9_{00}}(3m_{20}m_{23}^2 \!-\! 2m_{11}m_{23}m_{32 }\!-\! 4m_{02}m_{41}m_{23 }+ 3m_{02}m_{32}^2 \!-\! 4m_{14}m_{20}m_{32 }\\&+ m_{02}m_{14}m_{50 }- m_{05}m_{11}m_{50 }+ m_{05}m_{20}m_{41 }+ 3m_{11}m_{14}m_{41})
	\end{split}
	\]

\end{itemize}

 \section{Conclusion}
 In this paper, a novel and systematic scheme to determine moment invariants in n-dimensional space has been proposed.  
 More precisely, affine invariants are written under the form of linear combinations of invariants to non-uniform scale.
The coefficient vectors  of those linear combinations are obtained as null space of some matrices of integers. 
This  ensures that  all possible invariants from a moment vector are obtained. It also allows to remove 
 easily redundancy between invariants  using matrix algebra.

\end{document}